\documentclass[11pt,a4paper]{article}
\usepackage[hyperref]{acl2020}
\usepackage{times}
\usepackage{latexsym}

\usepackage{graphicx}
\usepackage{tabularx}
\usepackage{soul}

\usepackage{url}

\usepackage{epstopdf}
\usepackage[latin1]{inputenc}

\usepackage{hyperref}
\usepackage{xstring}
\usepackage{enumitem} 

\usepackage{cleveref}

\usepackage{xcolor}
\usepackage{epstopdf}

\usepackage{microtype}

\aclfinalcopy 

\title{ 
Practical and Ethical  Considerations in the Effective use of\\ Emotion and Sentiment Lexicons
}

\author{Saif M. Mohammad\\
National Research Council Canada\\
Ottawa, Canada\\ 
\texttt{saif.mohammad@nrc-cnrc.gc.ca.}}

\begin{document}
\maketitle
\begin{abstract}
Lexicons of word--emotion associations are widely used in research and real-world applications. As part of my research, I have created several such lexicons (e.g., the {\it NRC Emotion Lexicon}).
This paper outlines some practical and ethical considerations involved in the effective use of these lexical  resources.
\end{abstract}



\vspace*{10mm} 
\noindent \textbf{1. INTRODUCTION}\\[-11pt]

\noindent Words often convey affect (emotions, sentiment, feelings, and attitudes); either explicitly through their core meaning (denotation) or implicitly through connotation. 
For example, \textit{dejected} denotes sadness. 
On the other hand, \textit{failure} simply connotes sadness. 
Either through denotation or connotation, both words are associated with sadness. 
A compilation of such associations is referred to as a {\it word--affect association lexicon} (aka \textit{emotion lexicon}, aka \textit{sentiment lexicon}).
An entry in a lexicon usually includes a word, an emotion category or affect dimension (e.g., joy, fear, valence, arousal, etc.), and a score indicating association (or strength of association). 
The lexicons have 
a wide range of applications in commerce, public health, and research (not just in Natural Language Processing but also in Psychology, Social Sciences, Digital Humanities, etc.). 
Some notable examples include: tracking brand and product perception via social media posts, tracking support for controversial issues and policies, tracking buy-in for non-pharmaceutical health measures such as social distancing during a pandemic, literary analysis, and developing more natural dialogue systems. 

As part of my research, I have created and shared a number of word--emotion association lexicons  \cite{MohammadDD09,MohammadT10,Mohammad11a,Mohammad12,MohammadT13,MohammadSemEval2013,MohammadK14,Kiritchenko2014,arabicSA2015,SCL-NMA2016,OPP-lrec,SemEval2016Task7,vad-acl2018,LREC18-AIL}.\footnote{Homepage for Emotion lexicons: http://saifmohammad.com/WebPages/lexicons.html} 
Some of these lexicons such as the \textit{NRC Emotion Lexicon} \cite{MohammadT10,MohammadT13} and the \textit{NRC Valence, Arousal, and Dominance (VAD) Lexicon} \cite{vad-acl2018} were manually curated by crowdsourcing the task for tens of thousands of English words.\footnote{All of the manual annotation studies were approved by the NRC Research Ethics Board (NRC-REB). Protocol numbers can be found on corresponding project pages. REB review seeks to ensure that research projects involving humans as participants meet Canadian standards of ethics.} 
Some lexicons were generated automatically from large text corpora using simple statistical algorithms. 
Below I list some of the practical and ethical  considerations involved in the effective use of the lexicons.
Many of these considerations likely apply to any emotion lexicon (and not just to the ones I have created); 
for example, the General Inquirer \cite{Stone66}, ANEW \cite{nielsen2011new,bradley1999affective}, LIWC \cite{pennebaker2001linguistic},
MPQA \cite{Wiebe05}, and the lexicon by \newcite{warriner2013norms}.\\[-3pt]

\noindent \textbf{2. CONSIDERATIONS}\\[-11pt]

\noindent \textbf{a. Coverage:} Some lexicons have a few hundred terms, and some have tens of thousands of terms. However, even the largest lexicons do not include all the terms in a language. 
Mostly, they include entries for the canonical forms (lemmas), but some also include morphological variants. 
The high-coverage lexicons, such as the NRC Emotion Lexicon and the NRC VAD Lexicon have most common English words. 
However, when using the lexicons in specialized domains, one may find that a number of common terms in the domain are not listed in the lexicons. \\[-8pt]

\noindent \textbf{b. Dominant Sense Priors:} Words when used in different senses and contexts may be associated with different emotions. The entries in the emotion lexicons are mostly indicative of the emotions associated with the predominant senses of the words. This is usually not too problematic because most words have a highly dominant main sense (which occurs much more frequently than the other senses). 
In specialized domains, some terms might have a different dominant sense than in general usage. Entries in the lexicon for such terms should be appropriately updated or removed.
\\[-8pt]

\noindent \textbf{c. Associations/Connotations (not Denotations):}  A word that denotes an emotion is also associated with that emotion, but a word that is associated with an emotion does not necessarily denote that emotion. For example, \textit{party} is associated with joy, but it does not mean (denote) joy. The lexicons capture emotion \textit{associations}. Some have referred to such associations as \textit{connotations or implicit emotions}.\\[-8pt]

\noindent \textbf{d. Not Immutable:}
The associations do not indicate an inherent unchangeable attribute. 
Emotion associations can change with time, but these lexicon entries are largely fixed. They pertain to the time they are created or the time associated with the corpus from which they are created.\\[-8pt]

\noindent \textbf{e. Perceptions of emotion associations (not ``right" or ``correct" associations):} A useful question to ask before annotating language data is whether we are looking for ``correct" answers/labels or we want the annotators to tell us about how speakers of a language currently perceive the language data? For example, do I want somebody with academic training and expertise to annotate the data, or do I want to simply ask a large number of speakers of a language to get a sense of how the language data is perceived. For the lexicons that I have created, such as the NRC Emotion Lexicon and the NRC VAD Lexicon, our goal was explicitly to determine how speakers of a language perceive the emotion associations of words. Thus, the lexicons were created by crowdsourcing. We also specifically asked the annotators for `how speakers of a language perceive the word' as apposed to `the emotions evoked in the annotator'. The former led to less variance in responses (higher inter-annotator agreement) \cite{MohammadT13}.\\[-8pt]

\noindent \textbf{f. Socio-Cultural Biases:}
Since the emotion lexicons have been created by people (directly through crowdsourcing or indirectly through the texts written by people) they capture various human biases.
These biases may be systematically different for different socio-cultural groups. See papers describing individual lexicons to determine who produced the data (people from which countries, what is the gender distribution, age distribution, etc.). An advantage of crowdsourcing is that we obtain annotations from a wider pool of annotators; however, crowd annotators are systematically different from, and not representative of, the general population. \\[-8pt]

\noindent \textbf{g. Limitations of Aggregation by Majority Vote:} Each instance in a lexicon (word) is usually annotated by a number of annotators. Standard practice in aggregating the responses from multiple annotators is to take the most frequent response. However, it should be noted that sometimes other responses are also appropriate. Further, as noted above, different socio-cultural groups can perceive language differently, and taking the majority vote can have the effect of only considering the perceptions of the majority group. When these views are crystalized in the form of a lexicon, it can lead to the false perception that the norms so captured are ``standard" or ``correct", whereas other associations are ``non-standard" or ``incorrect". Thus, it is worth explicitly disavowing that view and stating that the lexicon simply captures the perceptions of the majority group among the annotators. I have made available the full set of disaggregated annotations, where possible. This can facilitate further work on teasing out different appropriate associations.\footnote{In case of the NRC Emotion Lexicon, the original annotations are at word-sense level. We created the word-level lexicon by keeping all majority-voted emotion associations for each of the word's senses.} 
Note that it is also problematic to consider 
all annotator responses as valid because sometimes annotators make mistakes, and some may have inappropriate biases (see next bullet).\\[-8pt]

\noindent \textbf{h. Inappropriate Biases:} Some of the human biases that have percolated into the lexicons may be rather inappropriate. For example, entries with low valence scores for certain demographic groups or social categories. 
In some instances, it can be tricky to determine whether the biases are appropriate or inappropriate. Capturing the inappropriate biases in the lexicon can be useful to show and address some of the historical inequities that have plagued humankind. Nonetheless, when these lexicons are used in specific tasks, care must be taken to ensure that inappropriate biases are not amplified or perpetuated. If required, remove entries from the lexicons where necessary.\\[-8pt]

\noindent \textbf{i. Source Errors:}
Even though the researchers take several measures to ensure high-quality and reliable data annotation (e.g.,
multiple annotators, clear and concise questionnaires, framing tasks as comparative annotations,
interspersed check questions, etc.), human-error can never be fully eliminated in large-scale annotations.
Expect a small number of clearly wrong entries.
Automatically generated lexicons also can have erroneous entries. They are often built on the assumption that the tendency of a word to co-occur with emotion-associated seed terms is proportional to its association with that emotion. However, in any corpus, there will always be some amount of chance high co-occurrences that are not accurate reflections of the true associations.\\[-8pt]

\noindent \textbf{j. Errors in Translation:}
Language resources are much more common in some languages (such as English) than in most other languages.
Thus often automatic translations of English resources are provided. For example,
the automatic translations of the NRC Emotion Lexicon and the VAD Lexicon are provided in over 100 languages. However, automatic translations can have errors; and the number of errors can vary depending on the language pair. Also, as noted earlier, there can be cultural differences in the emotion associations of a concept.
That said, several studies have shown that most emotion associations are fairly consistent across many language pairs \cite{redondo2007spanish, moors2013norms, schmidtke2014angst, mohammadSK2015, sianipar2016affective, yu2016building, stadthagen2017norms}.\\[-8pt]

\noindent \textbf{k. Relative (not Absolute):} The absolute values of the association scores themselves have no meaning. The scores help order the words relative to each other. For example, a term with a high valence score is associated with more positiveness than a term with with a lower valence score. 
Further, lexicons such as the VAD Lexicon that are created by comparative annotations (asking annotators to provide relative ordering among a small number of items at a time) do not claim that the mid-point valence score separates positive words from negative words. As stated earlier, the only claim is that terms with higher scores are associated with more (valence/arousal/dominance/etc.) than terms with lower scores.\\

\noindent \textbf{3. PRO-TIPS}\\[-5pt]

\noindent Here are some tips for effective and appropriate use of emotion lexicons:\\[-20pt]
\begin{enumerate}
\item Manually examine the emotion associations of the most frequent terms in your data. Remove entries from the lexicon that are not suitable (due to mismatch of sense, inappropriate human bias, etc.).\\[-18pt]
\item Depending on your specific use case, you may choose to re-scale the scores from 0 to 1 to -1 to 1, 1 to 10, etc. Note that if using the lexicon entries as features in machine learning experiments, the scale (0 to 1 or -1 to 1) can make a difference---e.g. in terms of how much weight should be assigned to terms with scores close to 0.\\[-18pt]
\item For text analysis, one can calculate various metrics such as the percentage of emotion words (when the lexicons provides a list of words associated with a category) or average emotion intensity (for real-valued associations).  \\[-18pt]
\item When the lexicons provide real-valued association scores (say between 0 and 1 for valence), one can calculate average valence of the words in the target text OR separate scores indicating average score of high-valence (positive) words and average score of low-valence (negative) words. You can use `$<0.33$' or `$>0.67$' as rough thresholds to determine low- and high-valence words.\footnote{Other thresholds such as `$<0.2$' and `$>0.8$', or `$<0.5$' and `$>0.5$' are also suitable. One may even determine these threshold by manually examining the lexicon entries.}  \\[-16pt]
\item If you are drawing inferences from texts using counts of emotion words:\\[-16pt]
\begin{enumerate}
\item It is more appropriate to make claims about emotion word usage rather than emotions of the speakers. For example, {\it `the use of anger words grew by 20\%'} rather than {\it `anger grew by 20\%'}. 
A marked increase in anger words is likely an indication that anger increased, but there is no evidence that anger increased by 20\%.\\[-14pt]
\item Comparative analysis is your friend. Often, emotion word counts on their own are not very useful. 
For example, {\it `the use of anger words grew by 20\% when compared to [data from last year, data from a different person, etc.]'} is more useful than saying {\it `on average, 5 anger words were used in every 100 words'}.
\item Inferences drawn from lager amounts of text are often more reliable than those drawn from small amounts of text.
 For example, {\it `the use of anger words grew by 20\%'} is informative when determined from hundreds, thousands, tens of thousands, or more instances. Do not draw inferences about a single sentence or utterance from the emotion associations of its constituent words. 
\end{enumerate}
\end{enumerate}
\noindent See these papers for examples of how some of these lexicons can be used \cite{mohammad-2011-upon, fraser-etal-2019-feel, hipson-mohammad-2020-poki, mendelsohn2020framework}. 
See proceedings of resent shared tasks on emotions such as SemEval-2018 Task 1: Affect in Tweets \cite{SemEval2018Task1} for how emotion lexicons are used in combination with training data and the latest machine learning techniques for emotion recognition.\footnote{https://competitions.codalab.org/competitions/17751}
You may also be interested in the applications compiled in this blog post: \href{https://medium.com/@nlpscholar/ten-years-of-the-nrc-word-emotion-association-lexicon-eaa47a8dd03e}{Ten Years of the NRC Word-Emotion Association Lexicon}.\footnote{https://medium.com/@nlpscholar/ten-years-of-the-nrc-word-emotion-association-lexicon-eaa47a8dd03e}\\ 

\noindent \textbf{4. CONCLUDING REMARKS}\\[-12pt]

\noindent Emotion lexicons can be simple yet powerful tools to analyze text. However, use of the lexicons (even for tasks that it is suited for) can lead to inappropriate bias. Applying a lexicon to any new data should only be done after first investigating its suitability, and requires careful analysis to minimize unintentional harm. I listed some considerations above that can help mitigate such unwanted outcomes. 
However, these are not meant to be comprehensive, but rather a jumping off point for further thought. The author welcomes feedback; including additional points to consider and include in this document. 
See also the papers associated with the lexicons for information about how the lexicons were created, their intended use, and their provenance. There is detailed information about each dataset in the research paper introducing it (often spanning 4 to 8 pages of the article).  See \citet{mohammad2020sentiment} (Section 10) 
for a brief discussion of the broad societal impacts of emotion and sentiment analysis. 

\bibliography{Ethics-Data-Statement-Emotion-Lexicons}
\bibliographystyle{acl_natbib}

\end{document}